\documentclass[a4paper,twoside]{article}

\usepackage{epsfig}
\usepackage{subcaption}
\usepackage{calc}
\usepackage{amssymb}
\usepackage{amstext}
\usepackage{amsmath}
\usepackage{amsthm}
\usepackage{multicol}
\usepackage{pslatex}
\usepackage{graphicx}
\usepackage{algorithm2e}
\usepackage{booktabs}
\usepackage{xcolor}
\usepackage{soul}
\usepackage{natbib}

\usepackage[bottom]{footmisc}
\usepackage{SCITEPRESS}     

\begin{document}

\title{Depth Estimation using Weighted-loss and Transfer Learning}

\author{\authorname{Muhammad Adeel Hafeez\sup{1}, Michael G. Madden\sup{1,2},Ganesh Sistu\sup{3} and Ihsan Ullah\sup{1, 2}}
\affiliation{\sup{1} Machine Learning Research Group, School of Computer Science, University of Galway, Ireland}
\affiliation{\sup{2}Insight SFI Research Centre for Data Analytics, University of Galway, Ireland}
\affiliation{\sup{3}Valeo Vision Systems, Tuam, Ireland}
\email{\{m.hafeez1, michael.madden, ihsan.ullah\}@universityofgalway.ie, ganesh.sistu@valeo.com} 
}

\keywords{Depth Estimation, Transfer Learning, Weighted-Loss Function}

\abstract {Depth estimation from 2D images is a common computer vision task that has applications in many fields including autonomous vehicles, scene understanding and robotics. The accuracy of a supervised depth estimation method mainly relies on the chosen loss function, the model architecture, quality of data and performance metrics. In this study, we propose a simplified and adaptable approach to improve depth estimation accuracy using transfer learning and an optimized loss function. The optimized loss function is a combination of weighted losses to which enhance robustness and generalization: Mean Absolute Error (MAE), Edge Loss and Structural Similarity Index (SSIM). We use a grid search and a random search method to find optimized weights for the losses, which leads to an improved model. We explore multiple encoder-decoder-based models including DenseNet121, DenseNet169, DenseNet201, and EfficientNet for the supervised depth estimation model on NYU Depth Dataset v2. We observe that the EfficientNet model, pre-trained on ImageNet for classification when used as an encoder, with a simple upsampling decoder, gives the best results in terms of RSME, REL and $log_{10}$: 0.386, 0.113 and 0.049, respectively. We also perform a qualitative analysis which illustrates that our model produces depth maps that closely resemble ground truth, even in cases where the ground truth is flawed. The results indicate significant improvements in accuracy and robustness, with EfficientNet being the most successful architecture. }

\onecolumn \maketitle \normalsize \setcounter{footnote}{0} \vfill

\section{\uppercase{Introduction}}
\label{sec:introduction}

In the context of computer vision, depth estimation is the task of finding the distance of different objects from the camera in an image. The process of depth estimation has been widely used in many application areas including Simultaneous Localization and Mapping (SLAM) \citep{alsadik2021simultaneous}, Object Recognition and Tracking \citep{yan2021depth}, 3D Scene Reconstruction \citep{murez2020atlas}, human activity analysis \citep{chen2013survey}, and more. 
\\

\vspace{-2mm}
Depth estimation can be done with various methods such as geometry-based methods in which the depth 3D information of an image is retrieved using multiple images captured from different positions. There are also sensor-based methods, which use LiDAR, RADAR and ultrasonic sensors for depth estimation. A single camera image can be post-processed by AI-based modalities for depth estimation, by leveraging advanced machine learning and computer vision techniques to estimate depth information from monocular images \citep{zhao2020monocular}. Sensor-based methods have multiple limitations such as hardware costs and high power requirements relative to camera-based methods, and are therefore often avoided in portable and mobile platforms \citep{sikder2021survey}. Although AI-based camera methods are more popular these days, they also have multiple limitations such as their computational cost, lack of interpretability and generalization challenges \citep{masoumian2022monocular}. \\
\vspace{-0.5mm}
Over the past few years, both unsupervised and supervised methods for depth estimation have become popular.  The unsupervised methods provide better generalization and reduce data annotation cost \citep{godard2017unsupervised}, whereas the supervised learning methods are more accurate and provide better explainability, in general, \citep{patil2022p3depth}. Supervised learning methods are typically characterized by their simplicity compared to unsupervised methods and the high degree of adaptability for future modifications and enhancements
 \citep{alhashim2018high}. 
 \\

 The accuracy of supervised methods usually depends upon two factors: 1) the loss function; 2) the model architecture. Based on the previous studies and experimental analysis, our goal in this paper is to propose a method which is simpler, easy to train, and easy and modify. For this, we mainly rely on optimizing existing loss functions and using transfer learning. Our experiments show that using high-performing pre-trained models via transfer learning, which were originally designed and trained for classification along with the optimized loss function can provide better accuracy, and reduce the root mean square error (RMSE) for the depth estimation problems.

\begin{figure*}[t]
  \centering
  \includegraphics[width=\textwidth]{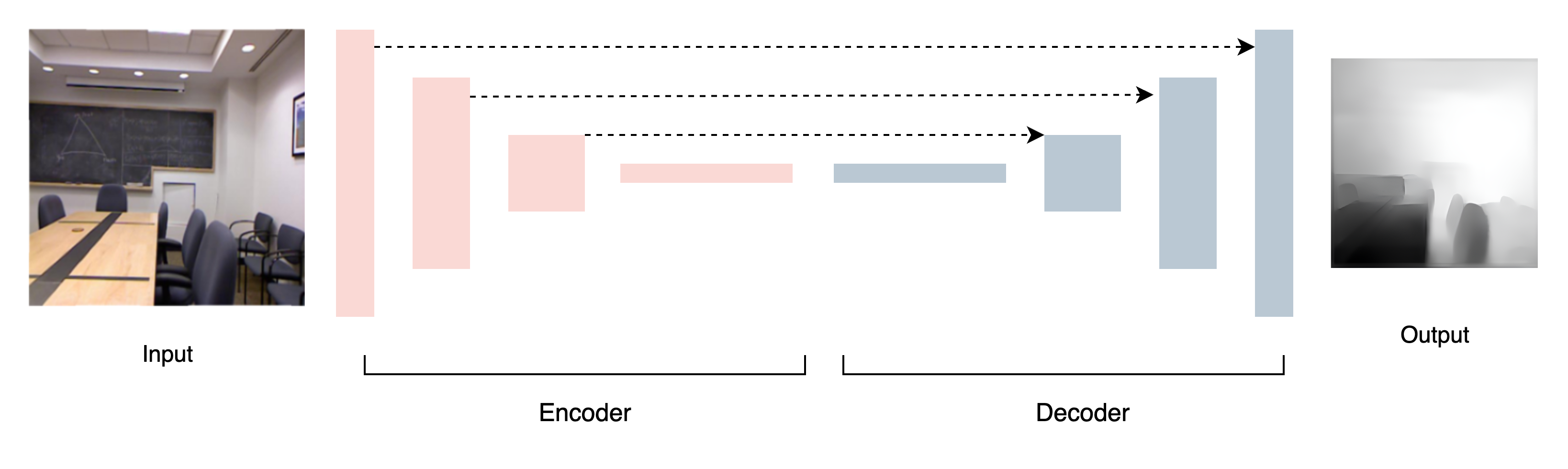}
  \caption{\textbf{Overview of the network.} We implemented a simple encoder-decoder-based network with skip connections. We changed the encoder between different models while keeping the decoder constant. The depth maps produced at the output were 1/2X of the ground-truth maps.}
  \label{fig:block-diagram}
\end{figure*}

Our main contributions are the following:
\begin{itemize}
    \item We propose an optimized loss function, which can be used for finetuning a pre-trained model.
    \item We perform an exploratory analysis with various pre-trained models. 
    \item After analysing the ground truth provided with datasets (NYU Depth Dataset v2), we identified some discrepancies in the dataset and how the proposed approach handled them. 
    \item We report the performance compared to the existing models and loss functions. 

    \end{itemize}
    
 While our approach advances the field of depth estimation, we acknowledge certain limitations in its current form, particularly for safety-critical applications where traditional image pair methods are renowned for their reliability \citep{mauri2021comparative}. 

\section{\uppercase{Literature Review}}

In the field of depth estimation, various methodologies have been explored over the years, including both traditional and deep learning-based approaches. Traditional depth estimation primarily relied on stereo vision and structured light techniques. Stereo vision methods, such as Semi-Global Matching (SGM) \citep{hirschmuller2008evaluation}, computed depth maps by matching corresponding points in stereo image pairs. Structured light approaches used known patterns projected onto scenes to infer depth \citep{furukawa2017depth}. These methods laid the foundation for depth estimation and remain relevant in specific scenarios.

In the past decade, %
deep learning-based depth estimation methods have made significant advancements. \citet{eigen2014depth} introduced an early 
deep learning model that used a convolutional neural network (CNN) to estimate depth from single RGB images. More recent supervised methods have introduced advanced architectures such as U-Net \citep{yang2021mixed} and MobileXNet \citep{dong2022mobilexnet} for improved depth prediction. 

Unsupervised depth estimation approaches have also gained prominence, eliminating the need for labelled data. \citet{garg2016unsupervised} proposed a novel framework that leveraged monocular stereo supervision, achieving competitive results without depth annotations. Other unsupervised methods use the concept of view synthesis, where images are reprojected from the estimated depth map to match the input images. This self-consistency check encourages the network to produce accurate depth maps without explicit supervision \citep{godard2017unsupervised}.

Traditional loss functions play a crucial role in training depth estimation models. Common loss functions include Mean Squared Error (MSE) \citep{torralba2002depth} and Mean Absolute Error (MAE) \citep{chai2014root}, which measure the squared and absolute differences between predicted and true depth values, respectively. Additionally, the Huber loss \citep{fu2018deep} offers a compromise between MSE and MAE, providing robustness to outliers. Huber loss is a hybrid loss function that uses MSE for small errors and MAE for large errors, making it more robust to outliers \citep{tang2019neural}. Other loss functions, such as structural similarity index (SSIM) \citep{wang2004image}, focus on perceptual quality, promoting visually enhanced depth maps. These losses have been reported individually \citep{carvalho2018regression} as well as in the combined functions, like MAE-SSIM, Edge-Depth, and Huber-Depth, enhancing the overall accuracy and perceptual quality of depth predictions \citep{paul2022edge}.
 
\section{\uppercase{Methodology}}
\label{sec:introduction}

In this section, we will discuss the dataset we used, the loss functions, the models, and the training process. 

\vspace{-3mm}

\subsection{Dataset}
In this study, we have used NYU Depth Dataset version 2 \citep{silberman2012indoor}. This dataset comprises video sequences from different indoor scenes, recorded by RGB and Depth cameras (Microsoft Kinect). The dataset contains 120,000 training images, with an original resolution of $640 \times 480$ for both the RGB and depth maps. In this dataset, the depth maps have an upper bound limit of 10 meters which means that any object that is 10 meters or more from the camera will have the maximum depth value. To reduce the decoder complexity and to save the training time, we kept the dimensions of the output of our model (depth maps) to half of the original dimensions ($320 \times 240$) and down-sampled the ground depth to the same dimensions before calculating the loss as reported previously \citep{li2018monocular}. The original data source contained both RAW (RGB, depth and accelerometer data) files and pre-processed data (missing depth pixels were recovered through post-processing). We used this processed data and did not apply any further pre-processing to the data. The test set contained 654 pairs of original RGB images and their corresponding depth values. 

\vspace{-2mm}
\subsection{Loss Functions}
\vspace{-2mm}
Loss functions play a crucial role while training a deep learning algorithm. Loss functions help to quantify the errors between the ground truth and the predicted images, hence enabling a model to optimize and improve its performance. In this study, we have used three different loss functions and combined them to get an overall loss. Details of all the individual losses are given in the sub-sections that follow.

\subsubsection{Mean Absolute Error (MAE)} 
Mean Absolute Error loss, also referred to point-wise loss, is a conventional loss function for many deep learning-based methods. It is the pixel-wise difference between the ground truth and the predicted depth and then the mean of these pixel-wise absolute differences across all pixels in the image.

It essentially quantifies how well a model predicts the depth at each pixel. MAE can be represented by the following equation:
\vspace{-2mm} 
\begin{equation}
L_{\text{MAE}} = \frac{1}{N} \sum_{i=1}^{N} |Y_{\text{true}_i} - Y_{\text{pred}_i}|
\end{equation}
Here N is the total number of data points or pixels in the image, $Y_{\text{true}_i}$ is the ground-truth depth of a pixel in the image while $Y_{\text{pred}_i}$ is the corresponding predicted depth of that pixel.

\vspace{-1mm}
\subsubsection{Gradient Edge Loss}
\vspace{-1mm}
Gradient edge loss or simply the edge loss calculates the mean absolute difference between the vertical and horizontal gradients of the true depth and predicted depth. This loss encourages the model to capture the depth transitions and edges accurately. The edge loss can be represented by the following equation. 
\vspace{-2mm} 
\begin{equation}
L_{\text{edges}} = \frac{1}{N} \sum_{I=1}^{N} \left(\left|\frac{\partial Y_{\text{pred}}}{\partial x} - \frac{\partial Y_{\text{true}}}{\partial x}\right| + \left|\frac{\partial Y_{\text{pred}}}{\partial y} - \frac{\partial Y_{\text{true}}}{\partial y}\right|\right)
\end{equation}

\noindent where $\frac{\partial Y}{\partial x}$ represent the horizantal edges, and $\frac{\partial Y}{\partial y}$ represent the vertical edges of the image Y. The edge loss helps to enhance the fine-grained spatial details in predicted depth maps. 

\subsubsection{Structural Similarity (SSIM) Loss}

This loss is used to compare the structural similarity between two images, and it helps to quantify how well the structural details are preserved in the predicted depth as compared to the true depth \citep{bakurov2022structural}. SSIM index can be represented by the following equation:

\begin{equation}
\resizebox{.50\hsize}{!}{
$\text{SSIM}(Y_{\text{pred}}, Y_{\text{true}}) = \frac{(2\mu_{Y_{\text{pred}}}\mu_{Y_{\text{true}}} + C_1)(2\sigma_{Y_{\text{pred}}Y_{\text{true}}} + C_2)}{(\mu_{Y_{\text{pred}}}^2 + \mu_{Y_{\text{true}}}^2 + C_1)(\sigma_{Y_{\text{pred}}^2} + \sigma_{Y_{\text{true}}^2} + C_2)}$
}
\end{equation}

In this equation, $\mu$ and $\sigma$ are the mean and standard deviation of the original depth maps and the predicted depth maps. $C$ terms are constants with small values to avoid any numerical instability in case $\mu$ or $\sigma$  are close to zero. The SSIM index between true depth and predicted depth ranges between -1 to 1. If the value of SSIM index is 1, it means that the depth maps are fully similar, otherwise, -1 indicates that depth maps are dissimilar. We converted this SSIM index into SSIM loss by simply subtracting the final value from 1 and scaling it with 0.5. This gives us the new range of SSIM loss which is between 0 to 1, where 0 means fully similar, and 1 means fully different depth maps.  This scaling is beneficial for the stability of gradient-based optimization algorithms used in training neural networks. 
 
\vspace{-2mm} 
\subsubsection{Combined Loss}
\vspace{-2mm} 
In this study, we have used a combined loss which is a  sum of weighted MAE, Edge Loss, and SSIM as reported in some previous studies \citep{alhashim2018high}. A combined loss promises to Enhance Robustness by addressing various challenges like fine details, edges and overall accuracy, as well as provide better generalization \citep{paul2022edge}. The combined loss function used in this study can be represented by the following:
\vspace{-2mm} 
\begin{equation}
    L_{\text{combined}} = w_1 \cdot L_{\text{SSIM}} + w_2 \cdot L_{\text{edges}} + w_3 \cdot L_{\text{MAE}}
\end{equation}

\noindent Here, $w_1$, $w_2$ and $w_3$ are the weights assigned to different losses. In previous studies \citep{alhashim2018high}, \citep{paul2022edge} authors have used these weights as 1, or 0.1 and no other values were explored or reported.  It is important to note that these three losses are somewhat independent of each other. The SSIM focuses on the structural similarities in depth maps, considering luminance, contrast, and structure. Edge loss targets the accuracy of edges, and MAE measures the pixel-wise absolute difference between the true and predicted depth. 

In this study, we explored that fine-tuning of the weights of the loss function is crucial and it directly affects the model's behaviour for the task of depth estimation. Adjusting the weights helps the model to adapt to the characteristics of the dataset and to be less sensitive to outliers, and it improves overall robustness. In order to find the optimized weights for the data, we used a grid search method and a random search method. For the grid search method, we initialized the weights to [0, 0.5, 1] and trained the model on a subset of the data. We made sure that this subset of the data should contain the maximum possible scenarios (Kitchen, Washroom, Living area etc) of the NYU2 data.  A combination of the weights which produced minimum validation loss during the training was kept, and the rest were discarded. 
We further explored the use of a random search instead of grid search. This time we used weight values of [0.2, 0.4, 0.6, 0.8, 1] and trained the model for 30 random combinations for a subset of the original data. We got the minimum validation loss for the following loss. 
\begin{equation}
    L_{combined} = 0.6 \cdot L_{MAE} + 0.2 \cdot L_{Edge} + L_{SSIM}
\end{equation}
We used this weighted loss function for the rest of the experiments. 

\subsection{Network Architecture}
In this study, we have used multiple encoder-decoder-based models for depth estimation using the NYU2 dataset. These models capture both global context and fine-grained details in depth maps, resulting in more accurate and visually coherent predictions. For the Encoder part, we used four different models: 
DenseNet121, DenseNet169, DenseNet201 and EfficientNet. All these models were pre-trained on ImageNet for classification tasks. These models were used to convert the input image into a feature vector, which was fed to a series of up-sampling layers, along with skip connections, which acted as a decoder and generated the depth maps at the output. We did not use any batch normalization or other advanced layers in the decoder, as suggested by a previous study \citep{fu2018deep}. Figure 1 shows a generic architecture of the network used in the study, where the encoder was changed with different state-of-the-art models as mentioned while the decoder was kept simple and constant. 

\subsection{Implementation and Evaluation}
To implement our proposed depth estimation network, we used TensorFlow and trained our models on two different machines, an Apple M2 Pro with 16GB memory and a machine with an NVIDIA GeForce 2080 Ti (4,352 CUDA cores, 11 GB of GDDR6 memory). The training time varied between machines and the model (for example, for DenseNet 169, it took 20 hours to train on GeForce 2080 Ti). The encoder weights were imported for different pre-trained models for classification on ImageNet and the last layers were fine-tuned. Decoder weights were initialized randomly, and in all experiments, we used the Adam optimizer with an initial learning rate of 0.0001.

\begin{figure}
  \centering
  \includegraphics[width=1\columnwidth]{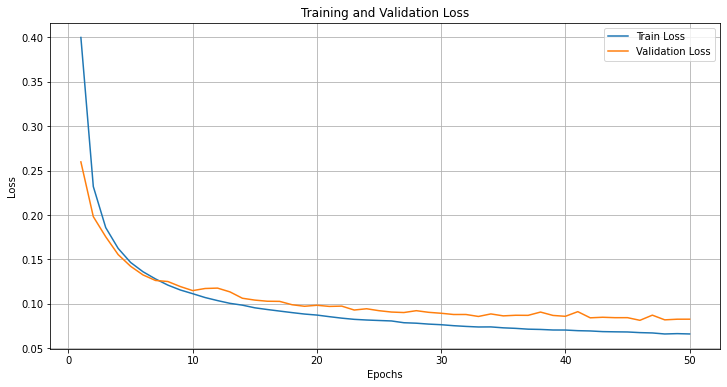}
  \caption{Training and Validation Loss for EfficientNet (50 epochs).}
  \label{fig:training_plot}
\end{figure}

Figure \ref{fig:training_plot} shows the training and validation loss for EfficientNet. The model was trained up to 50 epochs to be sure that the optimal stopping point considered for the training must contain a global minima for validation loss rather than local minima. Although the network still has the capacity to further train and converge, we adopted an early stopping approach (model weights from the 23rd epoch were taken) and will further explore this in future work. For the other models, we trained our network to 20 epochs to align with the existing research \citep{alhashim2018high}.  To evaluate our models, we used both quantitative and visual evaluations. 

\begin{figure*}[!t] 
  \vspace*{-0.7\baselineskip} 
  \centering
  \includegraphics[width=1\textwidth]{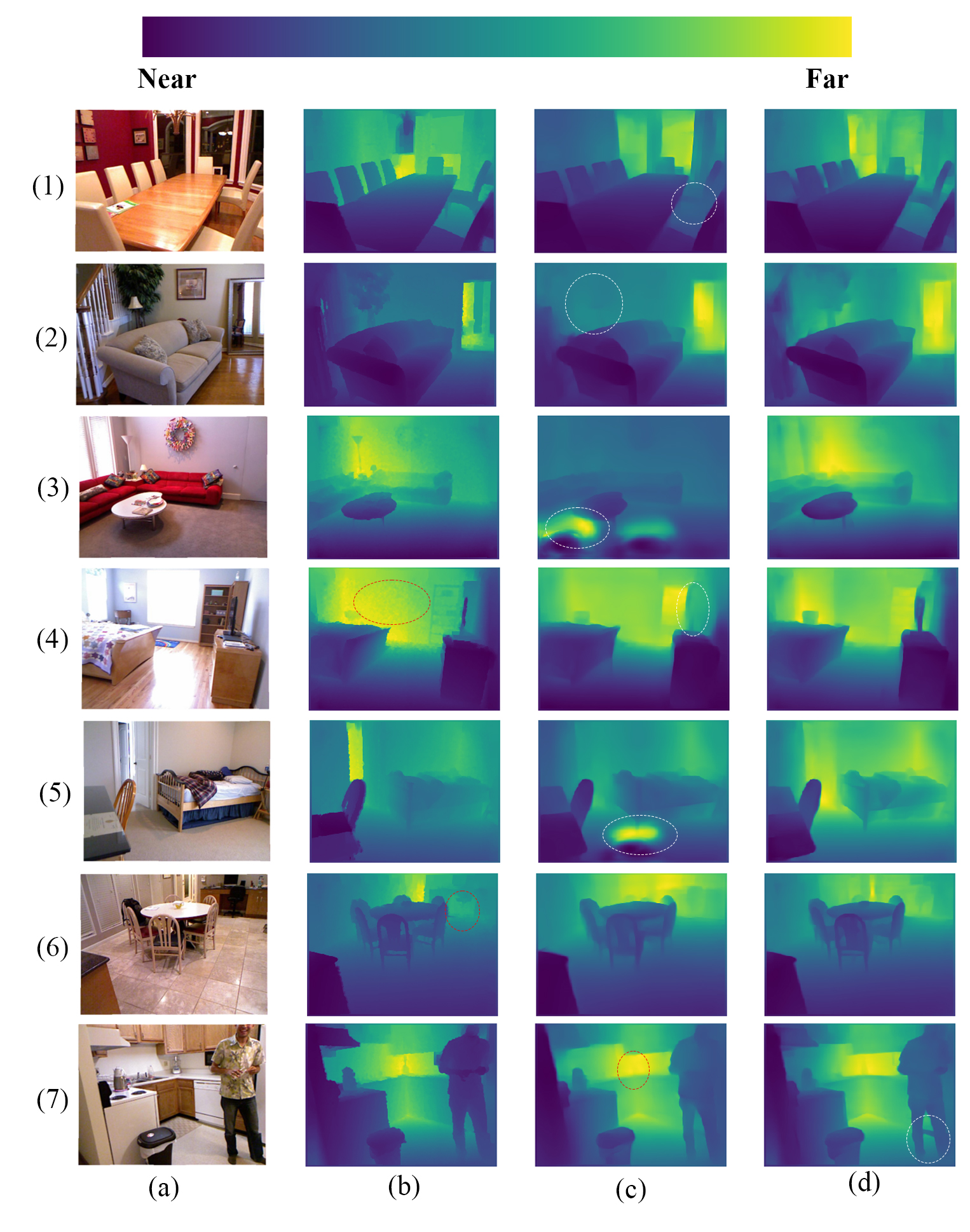} 
  \caption{The figure shows: (a) each original RGB image; (b) its ground-truth depth map; (c) the depth map predicted by DenseNet-169; (d) the depth map predicted by EfficientNet.}
  \label{fig:depthmaps}
\end{figure*}
\noindent\textbf{Quantitative Evaluation}: To compare the performance of the model with existing results in a quantitative manner, we used the standard six metrics reported in many previous studies \citep{zhao2020monocular}. These metrics include average relative error, root mean squared error, average $log_{10}$ error and threshold accuracies. The Relative Error (REL) quantifies the average percentage difference between predicted and true values, providing a measure of accuracy relative to the true values. It can be represented by the following formula:

\vspace{-2mm} 
\begin{equation}
\text{REL} = \frac{1}{N} \sum_{i=1}^{N} \frac{|Y_i - \widehat{Y}_i|}{Y}
\end{equation}

The Root Mean Squared Error (RMSE) can be defined as a measure of the average magnitude of the errors between predicted depth and true depth values and expressed as: 
\vspace{-2mm} 
\begin{equation}
    \text{RMSE} = \sqrt{\frac{1}{N} \sum_{i=1}^{N} (Y_i - \widehat{Y}_i)^2}
\end{equation}
The $log_{10}$ error measures the magnitude of errors between predicted and true values of depth on a logarithmic scale and is often used to assess orders of magnitude differences. 
\vspace{-2mm} 
\begin{equation}
    \text{log}_{10}\text{ error} = \log_{10}\left(\frac{1}{N} \sum_{i=1}^{N} \left|\frac{Y_i}{\widehat{Y}_i} - 1\right|\right)
\end{equation}

For all the above metrics, the lower values are considered more accurate. The last evaluation metric used in our study is threshold accuracy which is a measure that determines whether a prediction is considered accurate or not based on a specified threshold. It can be presented as:
\begin{equation}
    \text{TA} = \frac{1}{N} \sum_{i=1}^{N} \begin{cases} 1 & \text{if } |Y_i - Y^{\widehat{}}_i| \leq T \\ 0 & \text{otherwise} \end{cases}
\end{equation}

The threshold values we used are $T = 1.25, 1.25^2, 1.25^3$ which are commonly used in the literature i.e., \citep{godard2017unsupervised}.

\section{\uppercase{Results}}

In this section, we will discuss our experiment results based on the performance metrics discussed in the previous section, and we will also compare the results between different loss functions and CNN models. The purpose of all of these models was to predict depth maps. For a quantitative analysis,  we have defined the six performance metrics in section 3.4, where the increased threshold accuracy and decreased losses indicate a better-performing system.  Table 1 shows results obtained using optimized loss function for four different model architectures. This table indicates that using transfer learning on
pre-trained EfficientNet with optimized loss function outperformed all other models where the RMSE was reduced to 0.386. Comparing between different architectures of DenseNet, we found that the DenseNet169 gave the best results, compared with other architectures.  DenseNet201 was the third best model, whereas the DenseNet121 was the worst among these.

\begin{table}[h]
\centering
\caption{Comparison of different model architectures used as an encoder on the performance of Depth Estimation}
\scriptsize
\setlength{\tabcolsep}{8pt}
\begin{tabular}{lccccccc}
\toprule
\textbf{Model} & \textbf{$\delta_1\uparrow$} & \textbf{$\delta_2\uparrow$} & \textbf{$\delta_3\uparrow$} & \textbf{RMSE$\downarrow$} & \textbf{REL$\downarrow$} & \textbf{Log$_{10}\downarrow$} \\ 
\midrule
\textbf{DenseNet-121} & 0.812 & 0.936 & 0.951 & 0.587 & 0.137 & 0.059 \\
\textbf{DenseNet 169} & 0.854 & \textbf{0.980} & 0.994 & 0.403 & 0.120 & 0.047 \\
\textbf{DenseNet-201} & 0.844 & 0.969 & 0.993 & 0.501 & 0.123 & 0.052 \\
\textbf{EfficientNet} & \textbf{0.872} & 0.973 & \textbf{0.996} & \textbf{0.386} & \textbf{0.113} & \textbf{0.049} \\ 
\bottomrule
\end{tabular}
\end{table}

To provide a fair comparison, we have compared the performance of our model on similar studies on the NYU2 dataset. Table 2 provides a detailed comparison of our proposed model and the existing studies. For comparison purposes, we only took our best-performing model which is EfficientNet with the optimized loss function. The results show that EfficientNet along with the optimized loss function outperformed the existing approaches on different performance evaluators. 
\begin{table}[h]
\centering
\caption{Comparison of different model architectures used as an encoder on the performance of Depth Estimation}
\scriptsize
\setlength{\tabcolsep}{7pt}
\begin{tabular}{lccccccc}
\toprule
\textbf{Author} & \textbf{$\delta_1\uparrow$} & \textbf{$\delta_2\uparrow$} & \textbf{$\delta_3\uparrow$} & \textbf{RMSE$\downarrow$} & \textbf{REL$\downarrow$} & \textbf{Log$_{10}\downarrow$} \\
\midrule
\citep{laina2016deeper} & 0.811 & 0.953 & 0.988 & 0.573 & 0.127 & 0.055 \\
\citep{hao2018detail} & 0.841 & 0.966 & 0.991 & 0.555 & 0.127 & 0.042 \\
\citep{alhashim2018high} & 0.846 & \textbf{0.974} & 0.994 & 0.465 & 0.123 & 0.053 \\
\citep{yue2020edge} & 0.860 & 0.970 & 0.990 & 0.480 & 0.120 & 0.051 \\
\citep{paul2022edge} & 0.845 & 0.973 & 0.993 & 0.524 & 0.123 & 0.053 \\
\hline
\textbf{Ours} & \textbf{0.872} & 0.973 & \textbf{0.996} & \textbf{0.386} & \textbf{0.113} & \textbf{0.049} \\
\bottomrule
\end{tabular}
\end{table}

Figure 3 shows a brief qualitative comparison of results from two of the models we used in this study with the optimized loss function. Column (a) shows the real RGB images, whereas column (b) shows their ground truth depth maps as provided by the NYU2 dataset. Columns (c) and (d) shows the depth maps produced by DenseNet-169 and the EfficientNet respectively where the darker pixels correspond to near pixels and the brighter pixels represent the far pixels. The results show that the EfficientNet produced more coherent results where the depth maps are more close to the original depth maps. For example,  in Figure (1,c) the circled part shows that some portion of the chair, which was originally present in the ground truth depth was not recovered by DenseNet, but using the EfficientNet, in Figure (1,d) this part of the image was predicted in depth map very precisely. Similarly, figure (2,c) shows that the plant on the background which was present in the original image and ground truth depth was not properly predicted, but in Figure (2,d) the depth information about the plant is much better. In Figure (3,c) the DenseNet predicted wrong depth information, where the white circled portion was predicted as far pixel, which is not the actual case and the prediction was fine in case of (3,d).  Similarly, (5,c) also has some wrong depth prediction which was resolved in Figure (5,d). Besides this, we also observed some missing information in the ground truth depth maps. For example, in Figure (4,b) the background is a white wall, but in the ground-truth depth map, there is an incorrect noisy pattern. Figure (4,c) shows some missing information (depth perception of the TV) but in Figure (4,d) both of these issues were resolved. There is also a ground-truth error in Figure (6,b) in which the legs of the chair are missing, but our proposed model was able to reconstruct it from a single RGB image with much better accuracy. Our proposed model also made some wrong predictions for example, in (7,d) the circled area is in the background, as can be seen in (7,b) and (7,c), but our model predicted it as a near pixel.

\section{\uppercase{Conclusion}}

\textbf{Conclusion:} In this study, we have proposed a simple yet promising solution for depth estimation which is a common task in computer vision. Our primary aim was to enhance the quantitative and visual accuracy of depth estimation by investigating different loss functions and model architectures. For this purpose, we proposed an optimized loss function, which is the sum of three different weighted loss functions which are MAE, Edge loss and SSIM. We reported that chosen weights for the loss function, 0.6 for MAE, 0.2 for Edge Loss, and 1 for SSIM, consistently outperform other combinations. Additionally, we introduce a variety of encoder-decoder based models for depth estimation. Results showed that the EfficientNet pre-trained on ImageNet for classification task as encoder when used with a simple up-sampling decoder, and our optimized loss function gave the best results. To evaluate our proposed network, we have used both the qualitative methods (threshold accuracy, RMSE, REL and $log_{10}$ error) as well as visual or qualitative methods. \\
\noindent\textbf{Future Work:} In future work, we plan to enhance the reliability and interpretability of depth estimation for critical applications by adopting advanced statistical methods and explainable AI frameworks, inspired by \citep{bardozzo2022cross} we are planning to explore tools like Grad CAM and Grad CAM++ to provide clear insights into our model's decision-making process.  Furthermore, we will be using the attention mechanism for an even better visual representation of the depth maps and consider Pareto Front plots to further illustrate the various weight candidates for loss functions and how they impact the RMSE error on the validation set. 
\section*{\uppercase{Acknowledgements}}
This research is funded by Science Foundation Ireland under Grant number 18/CRT/6223. It is in partnership with VALEO.

\bibliographystyle{apalike}
{\small
\bibliography{example}}

\end{document}